\documentclass[a4paper, 10pt, conference]{ieeeconf}      

\IEEEoverridecommandlockouts                              

\usepackage{amsmath,nccmath,graphicx,epstopdf}
\usepackage[binary-units=true,per-mode=symbol]{siunitx}
\usepackage[colorinlistoftodos]{todonotes}
\usepackage{subfigure}

\title{\LARGE \bf
Steer with Me: A Predictive, Potential Field-Based Control Approach for Semi-Autonomous, Teleoperated Road Vehicles
}

\author{Andreas Schimpe and Frank Diermeyer 
\thanks{The authors are with the Institute for Automotive Technology at the Technical University of Munich (TUM), 85748 Garching bei M\"unchen, Germany. {\tt\small schimpe@ftm.mw.tum.de}}%
}

\usepackage{textcomp}
\usepackage{lipsum}
\newcommand\copyrighttext{%
\footnotesize \textcopyright 2020 IEEE. Personal use of this material is permitted. Permission from IEEE must be obtained for all other uses, in any current or future media, including reprinting/republishing this material for advertising or promotional purposes, creating new collective works, for resale or redistribution to servers or lists, or reuse of any copyrighted component of this work in other works.
}
\newcommand\copyrightnotice{%
    \begin{tikzpicture}[remember picture,overlay]
    \node[anchor=south,yshift=10pt, xshift=10pt] at (current page.south) {\fbox{\parbox{\dimexpr\textwidth-\fboxsep-\fboxrule\relax}{\copyrighttext}}};
    \end{tikzpicture}%
}

\begin{document}

\maketitle
\copyrightnotice
\thispagestyle{empty}
\pagestyle{empty}

\begin{abstract}
Autonomous driving is among the most promising of upcoming traffic safety technologies. Prototypes of autonomous vehicles are already being tested on public streets today. However, while current prototypes prove the feasibility of truly driverless cars, edge cases remain which necessitate falling back on human operators. Teleoperated driving is one solution that would allow a human to remotely control a vehicle via mobile radio networks. Removing in-vehicle drivers would thus allow current autonomous technologies to further progress towards becoming genuinely driverless systems. This paper proposes a new model predictive steering control scheme, specifically designed for semi-autonomous, teleoperated road vehicles. The controller is capable of receiving teleoperator steering commands and, in the case of potential collisions, automatically correcting these commands. Collision avoidance is incorporated into the design using potential fields. A term in the cost function facilitates natural maneuvers, and constraints on the maximum potential keep the vehicle at safe distances from obstacles. This paper also proposes the use of high-order ellipses as a method to accurately model rectangular obstacles in tight driving scenarios. Simulation results support the effectiveness of the proposed approach. 
\end{abstract}
%
%
\section{INTRODUCTION}
\label{CH:introSec1}
The development of fully automated and driverless vehicles has seen great effort in recent years, with great improvements being made in perception systems, planning, and control algorithms. To date, automated vehicles are already being tested with safety drivers on public streets. Assuming the automation detects if it cannot independently resolve a complex traffic situation, teleoperated driving can be leveraged as an enabling technology to smooth the transition towards truly driverless vehicles.  
\subsection{Teleoperated Driving}
\begin{figure}[t]
    \includegraphics[width=\columnwidth]{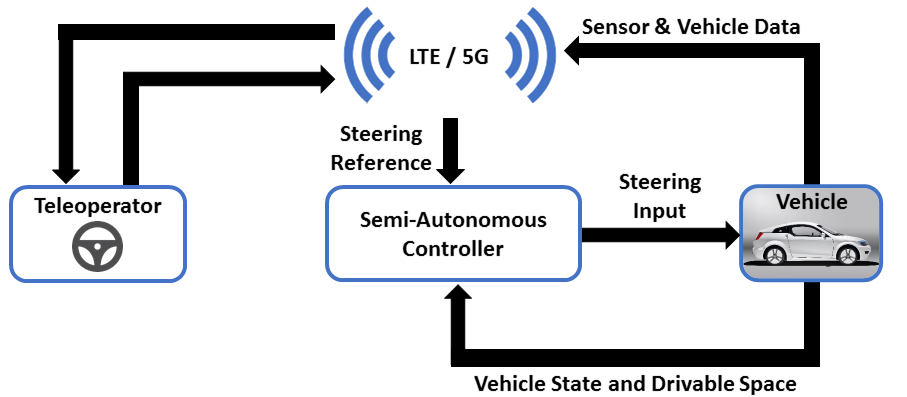}    
    \centering
    \caption{Control Architecture for the Teleoperated Vehicle}
    \label{fig:control_arch}
    \vspace{-3.5mm}
\end{figure}

For teleoperated driving, an interface is required to remotely control the vehicle. Through such an interface, sensor and vehicle data, e.g., video feeds and velocity, are transmitted via mobile radio networks from the vehicle to a remote control center. There, the data are displayed to the human teleoperator who generates control commands. These are then transmitted back to the vehicle for execution. \\
Teleoperated driving technology faces a number of challenges that need to be overcome. A system design for teleoperated road vehicles is presented in~\cite{Gnatzig2013}. A stable and secure mobile data connection between the vehicle and the control center needs to be established. This introduces latency, which can be critical if the vehicle is remotely controlled at stabilization level, i.e., the teleoperator produces direct steering commands. If the latency is too large, a different control concept has to be applied. In~\cite{Gnatzig2012}, a trajectory-based control scheme is proposed. The teleoperator plans a trajectory, which is transmitted to the vehicle for execution. A human-machine collaborative approach is presented in~\cite{Hosseini2014}. The automated driving function of the vehicle computes and suggests clusters of safe paths to the teleoperator, who then carries out the decision-making process by selecting one of them. \\
Reduced situational awareness poses one of the greatest challenges for teleoperated driving, as the teleoperator is not physically located in the vehicle. Consequently, additional mental effort is required to compensate for distortions and recreate missing information from the sensor data~\cite{Chelalli2011}. A teleoperated driving study, applying the direct control concept, is carried out in~\cite{Georg2018}. The use of head-mounted displays is analyzed and compared with conventional computer screens, so as to gauge the effect of different displays on situational awareness. For tight driving scenarios, the analysis shows that obstacles were hit with both interface designs. This leads to the use case and concept of the semi-autonomous control approach that is proposed in this paper, shown in Fig.~\ref{fig:control_arch}. It is assumed that an automated vehicle cannot always plan a trajectory that satisfies constraints of the vehicle kinematics, the road, and traffic regulations. When this occurs, the vehicle comes to a safe stop. The teleoperator takes over direct control by setting a desired speed and producing steering wheel angle commands. Onboard sensors continue to sense the drivable space around the vehicle. This enables the semi-autonomous controller to correct the steering input given by the teleoperator if the vehicle is at risk of hitting an obstacle.
%
%
\subsection{Related Work}
The approach is based on model predictive control~(MPC). In~\cite{Borrelli2005}, a first automotive application of MPC, tracking the desired trajectory, is presented. Following the idea from~\cite{Lam2010}, model predictive contouring control is applied for autonomous racing of RC cars in~\cite{Liniger2014}. The approach also incorporates avoidance of other opponents into the control problem. This work demonstrates the trend of tackling MPC problems with greater complexity and variety of objectives. For instance, a legible MPC design is proposed in~\cite{Bruedigam2018lmpc}. The framework aims at improving predictability of the planned maneuvers of the controlled vehicle in order to improve traffic flow on highways, while simultaneously optimizing for the passenger's comfort and energy efficiency. \\
For the control of semi-autonomous ground vehicles, a number of approaches can be found in literature. One such application is in advanced driver assistance functions, such as the predictive control framework presented in~\cite{Weiskircher2015}. Therein, a threat assessment based on the human driver's control inputs is performed by checking if motion constraints are violated. Controller intervention is then computed with the objective of minimizing the deviation of an output reference path. However, if this does not reflect the driver's intentions, the interventions of the controller may be deemed undesirable. In contrast, minimally invasive designs leave more freedom to the driver. In~\cite{Gray2013}, a predictive control framework for semi-autonomous vehicles with an uncertain, stochastic driver model is proposed. If necessary, the controller corrects the driver's steering with a minimal corrective steering action in order to satisfy lane departure constraints. Another complete framework for semi-autonomous control of ground vehicles with minimal restrictions is presented in~\cite{Anderson2012}. Similar to the previously mentioned approach, the controller enforces constraints in a desired homotopy. In~\cite{Erlien2016}, controlled and collision-free motion of the vehicle is ensured by using two so-called safe driving envelopes. One imposes constraints on the handling limits of the vehicle. The other formulates spatial limitations imposed by lane boundaries and obstacles. The controller only intervenes if a safe vehicle trajectory, lying within the envelopes, cannot be predicted based on the current driver's input. \\
All aforementioned approaches incorporate collision avoidance through constraints of spatial nature. This paper proposes the use of potential fields. In contrast, this enables more comfortable and natural maneuvers. For instance, a lane-associated potential field is introduced in the path planning MPC design for highway scenarios in~\cite{Liu2017}. A cost is calculated as a function of the longitudinal distance from the controlled vehicle to another vehicle in the same lane. On the one hand, the combination of MPC and potential fields yields the advantage of smooth planning with respect to the vehicle dynamics. On the other, it also enables the incorporation of different types of obstacles by using different potential functions. This is shown in~\cite{Rasekhipour2017}, which proposes a potential field-based MPC path-planning controller that utilizes crossable and noncrossable obstacles. 
\newpage
\subsection{Contributions}
This paper proposes a new model predictive steering control approach for semi-autonomous, teleoperated road vehicles. The primary objective is to track an input reference that is generated by the teleoperator. Secondly, the approach is meant to assist and correct the teleoperator's steering actions if the controlled vehicle is at risk of hitting an obstacle. In the proposed design, collision avoidance is incorporated using a potential field that models surrounding obstacles. On the one hand, this potential field is part of the cost function of the predictive control formulation. On the other, constraints are imposed on an uncrossable value of the potential field. Additionally, this paper proposes the use of ellipses with higher orders. As is shown, this improves the accuracy of bounding rectangular obstacles. Finally, the presented approach is validated in simulation with a simulated teleoperator. 
\section{MODELING}
\label{CH:modelSec2}

\subsection{Kinematic Bicycle  Model}
\label{CH:vehMdl}
In the MPC scheme, the motion of the vehicle is predicted using the kinematic bicycle model. This choice is motivated by only moderate driving speeds during the teleoperation, not necessitating sophisticated dynamic vehicle models. \\
Referring to~\cite{Kong2015}, the vehicle model equations of the center of mass~(CoM) are given by 
\begin{subequations} \begin{align}
\dot{x} &= v \, \cos( \theta + \beta), \label{EQ:dx} \\
\dot{y} &= v \, \sin(\theta + \beta), \label{EQ:dy} \\
\dot{\theta} &= \frac{v}{l_{\textrm{r}}} \, \sin(\beta), \label{EQ:dtheta} \\
\beta &= \arctan \left( \frac{l_{\textrm{r}}}{l_{\textrm{f}}+l_{\textrm{r}}} \, \tan(\delta) \right). \label{EQ:beta}
\end{align}
\label{EQ:mdlEqs}%
\end{subequations}
The location and orientation of the CoM in an inertial frame is represented by the coordinates~$x$ and~$y$, and the heading angle~$\theta$. The velocity is given by~$v$. The location of the CoM between the front and rear axle is given by the distances~$l_{\textrm{f}}$ and~$l_{\textrm{r}}$. The side-slip angle~$\beta$ describes the direction of movement of the CoM with respect to the longitudinal axis of the vehicle. The steering angle at the front wheel is given by~$\delta$. \\
To ensure a safe motion of the vehicle, the control design proposed in this paper incorporates collision avoidance for the front left~($\textrm{fl}$) and front right~($\textrm{fr}$) edge of the vehicle. These positions are calculated from the pose of the CoM by
\begin{subequations} \begin{align}
x_{\textrm{f}*} &= x + l_{\textrm{f}}' \, \cos(\theta) \mp \frac{w}{2}\, \sin(\theta), \\
y_{\textrm{f}*} &= y + l_{\textrm{f}}' \, \sin(\theta) \pm \frac{w}{2}\, \cos(\theta),
\end{align} \label{EQ:edges}%
\end{subequations}
with~$* \in \{ \textrm{l},\textrm{r} \}$. The distance from the CoM to the front bumper is~$l_{\textrm{f}}'$. The width of the vehicle is given by~$w$. 
\newpage
\subsection{Potential Field for Obstacle Avoidance}
\label{CH:potField}
In motion planning for mobile robots, a potential field~$P$, evaluated in the~$xy$-plane is generated by a sum over individual potential functions~$P_q$. This yields
\begin{equation}
P(x,y) = \sum_q P_q(x,y),
\end{equation}
with~$q$ denoting the index of the $q$th potential function. A potential function can either be attractive, to represent a goal region of the controller, or repulsive, to model obstacles or road structures. This paper presents a design, which uses the potential field to incorporate obstacle avoidance. Thus, the potential functions are repulsive only. They are given by 
\begin{equation}
P_q(x,y) = \frac{\alpha}{\left( s_q(x,y)+1 \right)^\beta},
\label{EQ:potFunc}
\end{equation}
with the parameters~$\alpha$ and~$\beta$ specifying the strength and slope of the potential function. The implicit function~$s_q$ describes the shape of the modeled obstacle. Where the value of~$s_q$ equals zero,~$P_q$ evaluates to~$\alpha$. This value should be non-crossable for the vehicle. Thus, it is incorporated as a constraint in the MPC formulation. \\
The use of the presented potential function formulation yields the advantage that the function~$s_q$ can be chosen to represent any obstacle shape. In order to facilitate the use of optimization algorithms such as sequential quadratic programming, the only requirement to be fulfilled is that~$s_q$ is twice continuously differentiable. Furthermore, compared to incorporating collision avoidance as spatial constraints for each obstacle, the potential field yields advantages in scalability. The number of obstacles to be avoided does not affect the number of constraints that need to be satisfied in the optimization problem. 
\subsection{High-Order Ellipses for Rectangular Obstacles}
\begin{figure}[t]
    \includegraphics[width=\columnwidth]{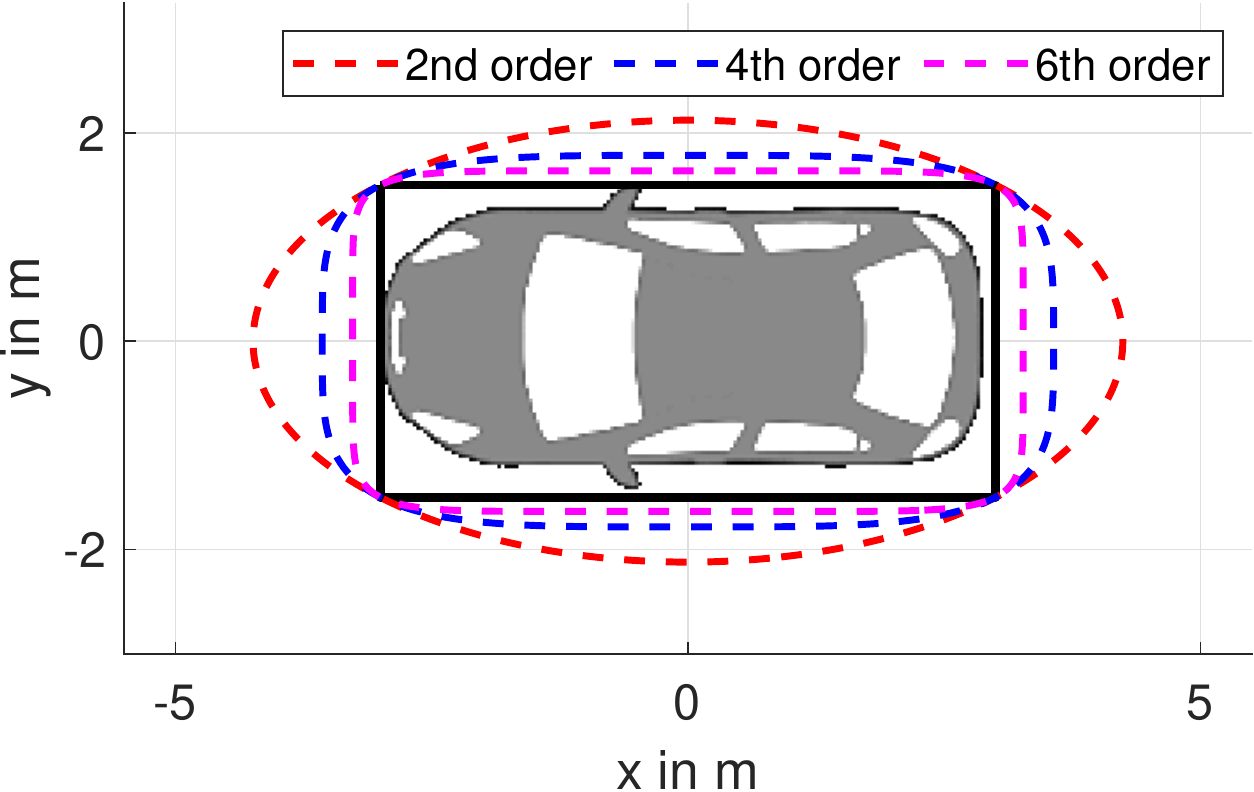}
    \centering
    \caption{Higher Order Ellipses for Approximating Rectangular Obstacles}
    \label{fig:ellipses}
    \vspace{-3.5mm}
\end{figure}

To incorporate collision avoidance for rectangular obstacles, e.g., other vehicles, an accurate analytic representation for the shape has to be found. For instance in~\cite{Andersen2017}, a vehicle is described as a union of four longitudinally shifted circles. This provides an accurate approximation of the vehicle shape, but also yields a higher computational burden than one single expression. A superior choice for the presented design is the shape of an ellipse. Omitting the index~$q$ for better readability, the shape function~$s$ is given by
\begin{equation}
s(x,y) = \left( \frac{x-x_\textrm{e}}{a} \right)^n + \left ( \frac{y-y_\textrm{e}}{b} \right )^n - 1,
\label{EQ:ellipse}
\end{equation}
with the center of the ellipse located at~$(x_\textrm{e},y_\textrm{e})$, the dimensions quantified by the semi axis~$a$ and~$b$, and the even order~$n$ of the ellipse. In literature, an order of~$2$ and relatively large dimensions are commonly chosen, generously constraining large areas around the obstacles. For instance in~\cite{Bruedigam2018smpc}, this is to improve safety in dynamic, autonomous highway scenarios. \\
This paper presents an approach for static obstacles, but in geometrically tight scenarios. Thus, the obstacles need to be modelled with higher accuracy. It is proposed to calculate the semi axis by equally scaling the length~$L$ and width~$W$ of the rectangular obstacle. This yields
\begin{equation}
a = f \cdot L/2 \quad \textrm{and} \quad b = f \cdot W/2.
\label{EQ:abEll}\end{equation}
The scaling factor~$f$ is chosen such that the ellipse just encloses the edge of the rectangle. Thus, it has to be
\begin{equation}
f = \sqrt[n]{2}.
\end{equation}
As indicated above, an ellipse order of~$n = 2$ does not satisfy the requirement of accurately approximating rectangular obstacles for the use case of the presented approach. Thus, this paper proposes to increase the order. Fig.~\ref{fig:ellipses} shows how different ellipse orders bound a rectangular obstacle. An order of~$2$ is a conservative choice, lavishly enclosing the obstacle and creating large buffer zones in longitudinal direction. Increasing the order yields higher accuracy, making this representation suitable for tight driving scenarios. 
%
%
%
%
\section{PREDICTIVE CONTROL FORMULATION}
\label{CH:mpcSec3}
To incorporate various control objectives and constraints, the control scheme in this paper is formulated as a model predictive control problem. In MPC, a constrained optimization problem is solved at each sampling instant~$t$ to determine a sequence of optimal control inputs over a discretized, finite time horizon, the prediction horizon. The first input of this sequence is applied to the system. With new measurements, the procedure is repeated at the next sampling instant. \\
In the following, the proposed MPC formulation is presented. The states and outputs of the system, the kinematic vehicle model, are denoted by~
$\xi \nolinebreak = \nolinebreak \begin{bmatrix} x ,~ y ,~ \theta \end{bmatrix}^\textrm{T}$
and~
$\eta \nolinebreak = \nolinebreak \begin{bmatrix} x_{\textrm{fl}} ,~ y_{\textrm{fl}} ,~ x_{\textrm{fr}} ,~ y_{\textrm{fr}} \end{bmatrix}^\textrm{T}$, respectively.
The input of the system is the steering angle $\delta$. The prediction horizon of the controller is~$N$. If this is used in the superscript, the symbol represents the sequence of the respective quantity over the prediction horizon. \\
The objectives of the controller are three-fold. These are matching the teleoperator's intentions, collision avoidance, and the smoothness of the control inputs. The first term of the cost function penalizes deviation of the controller input from the current teleoperator's input reference~$\delta_{\textrm{ref}}(t)$. With the weight~$w_{\delta_{\textrm{ref}}}$, the formulation reads
\begin{equation}
J_{\delta_{\textrm{ref}}} \left ( \delta_0 \right ) = w_{\delta_{\textrm{ref}}} \left ( \delta_0 - \delta_{\textrm{ref}}(t) \right )^2 .
\end{equation}
%
Penalizing only the first instant, and not the complete prediction horizon, avoids that the controller intervenes too early. However, it also results in sudden, large interventions, as the controller waits until the last possible moment to correct the steering. \\
The second cost function term is the potential field for collision avoidance, which pushes the vehicle away from obstacles. The potential field is evaluated at the system outputs, which are the front edges of the vehicle. These values are summed over the complete prediction horizon. With the weighting factor~$w_P$, the cost function term yields
\begin{equation}
J_P \left( \eta^N \right) = w_P \sum_{i=1}^{N} \Big ( P(x_{\textrm{fl},i},y_{\textrm{fl},i}) + P(x_{\textrm{fr},i},y_{\textrm{fr},i}) \Big ). 
\end{equation}
The third cost function term weights the predicted steering rates in order to improve smoothness. With the weighting factor~$w_{\dot{\delta}}$, it is given by
\begin{equation}
J_{\dot{\delta}} \left ( \delta^N \right ) = w_{\dot{\delta}} \sum_{i=1}^{N-1} (\delta_i - \delta_{i-1})^2.
\end{equation}
%
Altogether, the cost function of the controller is given by the sum of the three presented terms. With the constraints, the complete optimization problem reads
\begin{subequations}
\begin{equation}
\min_{\delta^N} J \left (\eta^N,\delta^N \right ) = J_{\delta_{\textrm{ref}}} \left (\delta_0 \right ) + J_P \left (\eta^N \right ) + J_{\dot{\delta}} \left (\delta^N \right )
\label{EQ:costFct}
\end{equation}
subject to
\begin{align}
\xi_{i+1} = f(\xi_{i}, \delta_{i}), \label{EQ:vehMdlConstr} \\
\eta_{i+1} = g(\xi_{i+1}), \label{EQ:vehEdgeConstr} \\
\xi_0 = \xi(t), \label{EQ:initStateCond} \\
\delta^{\textrm{min}} \leq \delta_i \leq \delta^{\textrm{max}}, \label{EQ:steerConstr} \\
\dot{\delta}^{\textrm{min}} \leq \dot{\delta}_i \leq \dot{\delta}^{\textrm{max}}, \label{EQ:steerRateConstr} \\
P(x_{\textrm{fl},i+1},y_{\textrm{fl},i+1}) \leq \alpha, \label{EQ:lPFCon}\\
P(x_{\textrm{fr},i+1},y_{\textrm{fr},i+1}) \leq \alpha, \label{EQ:rPFCon}\\
(i = 0,1,...~N-1). \notag
\end{align}
\label{EQ:mpcProb}%
\end{subequations}
The system model states~$f(\xi_i,\delta_i)$ are the kinematic vehicle model equations~\eqref{EQ:mdlEqs}, discretized by the time $t_d$ using Forward Euler. The system model outputs~$g(\xi_{i+1})$ are the two front edges that are calculated from~\eqref{EQ:edges}. The equality constraint~\eqref{EQ:initStateCond} formulates the initial state condition. Physical actuation constraints of the minimum and maximum steering magnitude~$( \delta^{\textrm{min}} , \delta^{\textrm{max}} )$ and rate~$( \dot{\delta}^{\textrm{min}} , \dot{\delta}^{\textrm{max}} )$ are incorporated by~\eqref{EQ:steerConstr} and~\eqref{EQ:steerRateConstr}. The steering rates are calculated using numerical differentiation. The cost function already incorporates collision avoidance in a soft manner by the objective of minimizing the values of the potential field. However, this alone does not guarantee safety. Thus, the constraints~\eqref{EQ:lPFCon} and~\eqref{EQ:rPFCon} are introduced. These impose an upper bound~$\alpha$, from~\eqref{EQ:potFunc}, on the potential values. As described earlier, this value is reached where the shape function~$s_q$ evaluates to zero, which must not be crossed by the vehicle. 
\section{SIMULATION RESULTS}
\label{CH:simsSec4}
Simulations were performed in MATLAB/Simulink\textsuperscript{{\textregistered}} to validate the proposed control approach. Following~\cite{Carvalho2013}, the optimization is carried out using an MPC-tailored sequential quadratic programming~(SQP) algorithm. The algorithm iteratively approximates the optimization problem about an operating point. From this, a quadratic programming~(QP) subproblem is created and solved. The solution is then used to update the operating point for the next iteration. This procedure is repeated at each sampling instant until convergence, or a maximum of three iterations is reached. In the first iteration, a reasonable initial guess of the operating point is obtained by shifting the solution of the previous sampling instant. The algorithm can be run online in less than~\SI{30}{\milli\second} on an Intel~i7~\SI{2.6}{\giga\hertz}~6~core processor with~\SI{16}{\giga\byte} of RAM. In this work, the QP subproblems are solved using the MATLAB\textsuperscript{{\textregistered}} Optimization Toolbox routine~\texttt{quadprog}, which makes up~\SI{45}{\percent} of the computation time on average. It is expected that the total computation time can be further reduced by utilizing a faster QP solver. The usage of ellipses with higher orders adds only minimal, neglectable computation time to the QP setup phase. \\
In the following, results of a parking lot and a lane change scenario are presented. Each scenario is run twice, once without (unassisted) and once with (assisted) the steering corrections of the proposed controller. The scenarios consist of a number of obstacles and a desired path. The desired path is only known to the teleoperator who is generating the steering input reference. The teleoperator is simulated using a feedback-linearized path-tracking controller, taken from~\cite{Burnett2019}. Omitting the time index, the control law is
\begin{equation}
\delta_{\textrm{FBL}} = \arctan \left( \frac{ - \gamma_1 \, e_{\textrm{L}} - \gamma_2 \, v \, \sin \left( e_{\textrm{H}} \right)}{v^2 \, \cos\left( e_{\textrm{H}} \right)} \right), 
\end{equation}
with the current vehicle velocity~$v$, lateral error~$e_{\textrm{L}}$, and heading error~$e_{\textrm{H}}$. Tracking behavior can be traded off with the controller gains~$\gamma_1$ and~$\gamma_2$. Another feedback term is introduced in order to simulate visual or haptic feedback that would be provided to a human teleoperator. This term is proportional to the deviation between the steering of the teleoperator and the controller at the previous sampling instant. With another gain~$\gamma_3$, the complete simulated teleoperator model yields
\begin{equation}
\delta_{\textrm{ref}}(t) = \delta_{\textrm{FBL}}(t) + \gamma_3 \left( \delta_0(t-t_\textrm{s}) - \delta_\textrm{FBL}(t) \right).
\end{equation}
The sampling time of the teleoperator and the predictive controller is~$t_s = \SI{50}{\milli\second}$. The discretization time between the instants of the prediction horizon~$N = 12$ is~$t_d = \SI{200}{\milli\second}$. This choice is motivated by the study on discretization effects of the kinematic vehicle model, presented in~\cite{Kong2015}. Therein, it was found that the prediction accuracy of the kinematic vehicle model improves with larger discretization times. The order of the ellipses that are enclosing the obstacles, is chosen to be~$n = 4$. The potential functions use the parameters~$\alpha = 1$ and~$\beta = 1$. This parametrization yields a potential field that only influences the controller in immediate proximity to obstacles. The steering inputs are constrained by
\begin{subequations} \begin{equation}
\delta_i \in [-35^{\circ} , 35^{\circ}],
\end{equation} \begin{equation}
\dot{\delta}_i \in [ \SI{-30}{\degree\per\second} , \SI{30}{\degree\per\second} ].
\end{equation}
\end{subequations}
During teleoperated driving, only low speed is considered. That is~\SI{3}{\meter\per\second} in the presented scenarios. The values of the controller gains and weights are reported in the appendix. 
\begin{figure}[t]
    \includegraphics[width=\columnwidth]{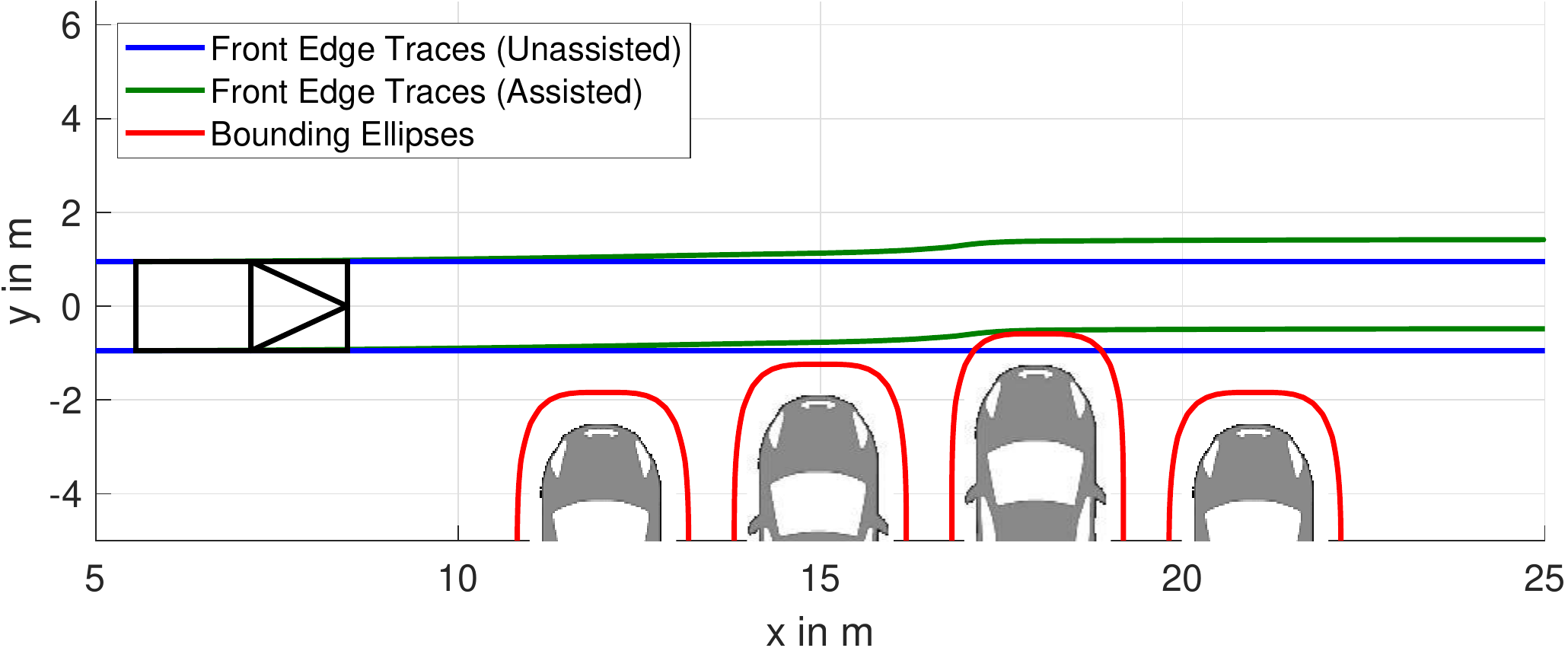}
    \centering
    \caption{Parking Lot Scenario}
    \label{fig:scenPVtraj}
	\vspace{-3.5mm}
\end{figure}
\begin{figure}[t]
\subfigure[Steering Profiles]
{
    \includegraphics[width=\columnwidth]{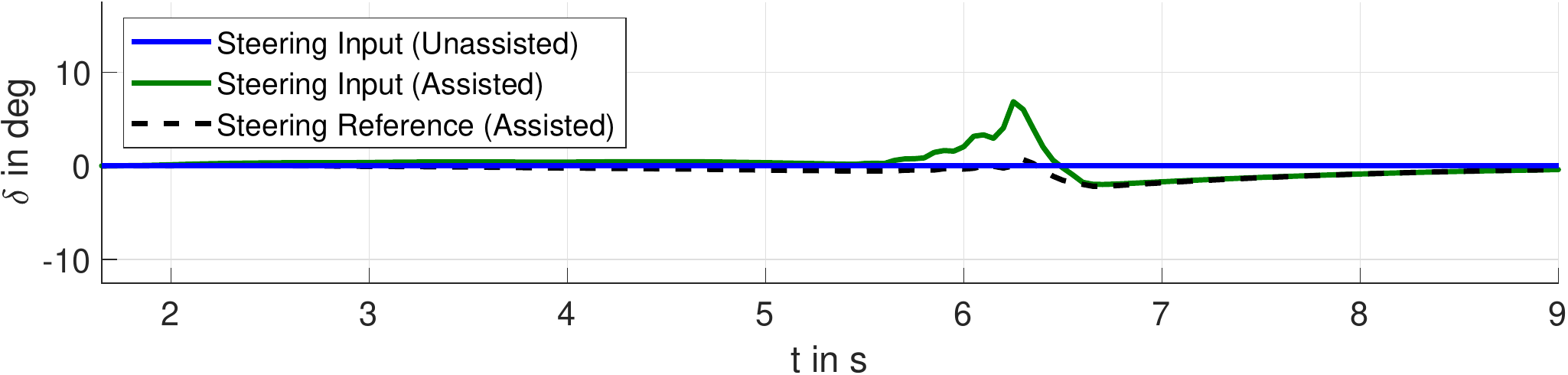}
    \label{fig:scenPVsteer}
}
\subfigure[Potential Field]
{
    \includegraphics[width=\columnwidth]{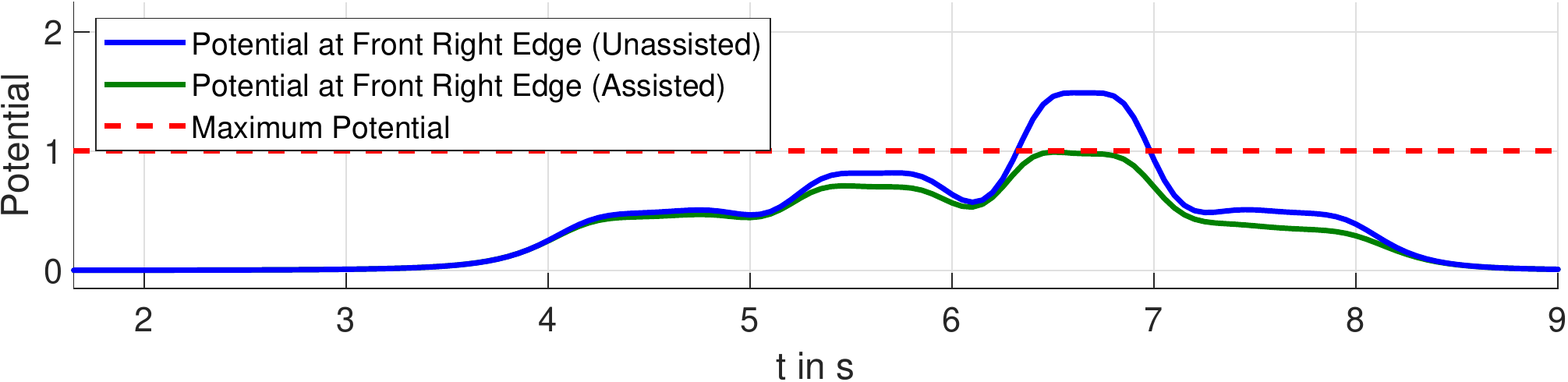}
    \label{fig:scenPVpotf}
}
\centering
\caption{Steering Profiles and Potential Field of the Parking Lot Scenario}
\label{fig:scenPVsteerNpotf}
\vspace{-3.5mm}
\end{figure}
\subsection{Parking Lot Scenario}
The first scenario takes place in a parking lot. As shown in Fig.~\ref{fig:scenPVtraj}, the teleoperator intends to pass four parked vehicles. The traces of the vehicle front edges are shown for the unassisted run~(blue), and the assisted run~(green). The teleoperator misjudges the distance to the parked vehicles. In the unassisted run, this leads to the vehicle coming too close to the third parked vehicle. The bounding ellipse~(red) is entered. With assistance, the controller corrects the steering to keep the vehicle safe. The gains of the feedback-linearized controller are set to only correct deviations of the desired heading, which is zero. Thus, the vehicle is steered back to continue straight after passing the obstacles. \\
The steering profiles of the two simulation runs are shown in Fig.~\ref{fig:scenPVsteer}. The potential field, evaluated at the right front edge, is shown in Fig.~\ref{fig:scenPVpotf}. In the unassisted run~(blue), the teleoperator's steering remains constant at zero, as there is no deviation of the desired heading. The safety constraint of the maximum potential~(red) is violated. With the assistance of the controller, the steering input applied to the vehicle~(green) deviates from the teleoperator's steering reference~(black) when passing the third obstacle. By this, the safety constraint of the maximum potential is satisfied. 
%
%
\begin{figure}[t]
    \includegraphics[width=\columnwidth]{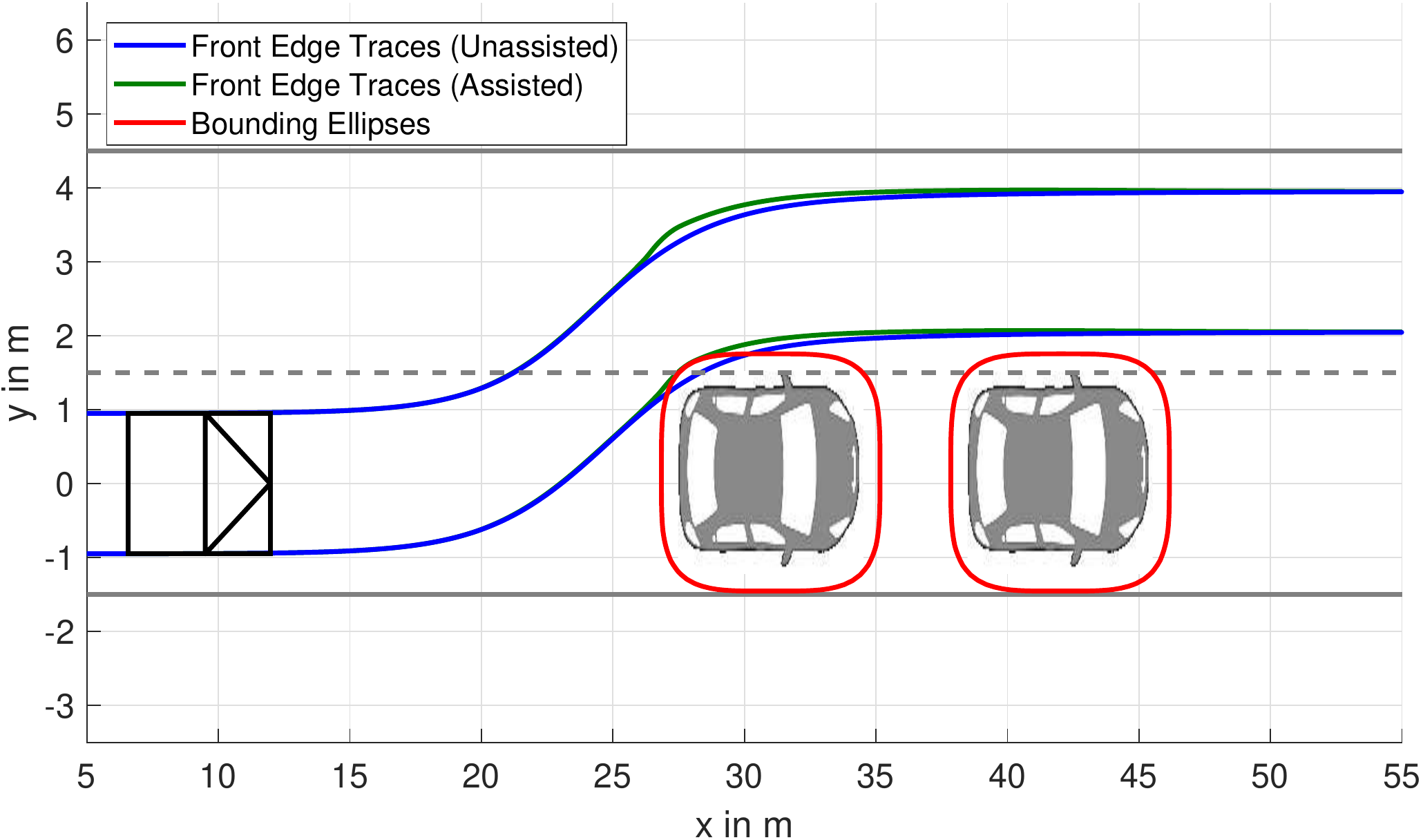}
    \centering
    \caption{Lane Change Scenario}
    \label{fig:scenLCtraj}
\end{figure}
\begin{figure}[t]
\subfigure[Steering Profiles]
{
    \includegraphics[width=\columnwidth]{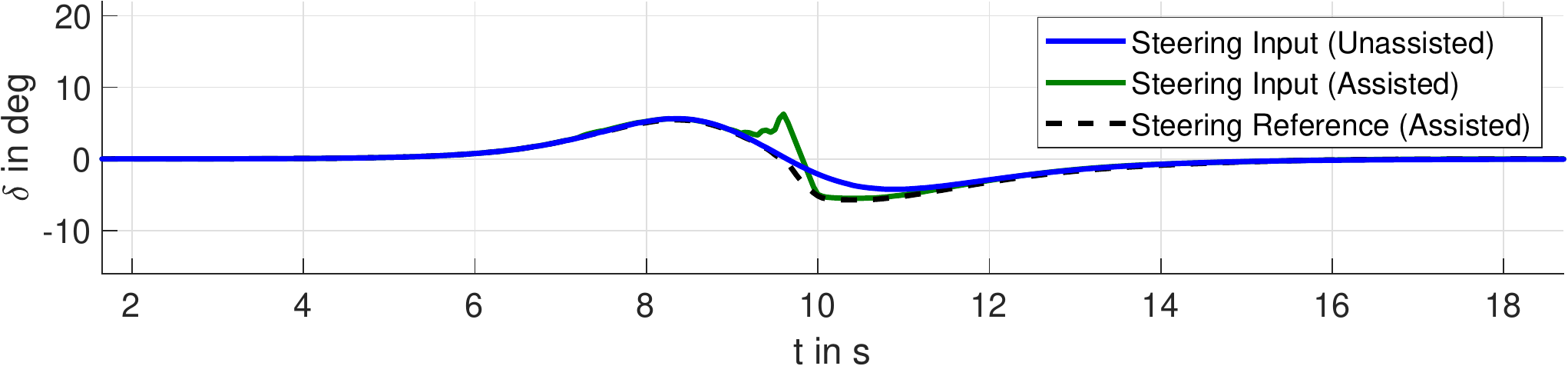}
    \label{fig:scenLCsteer}
}
\subfigure[Potential Field]
{
    \includegraphics[width=\columnwidth]{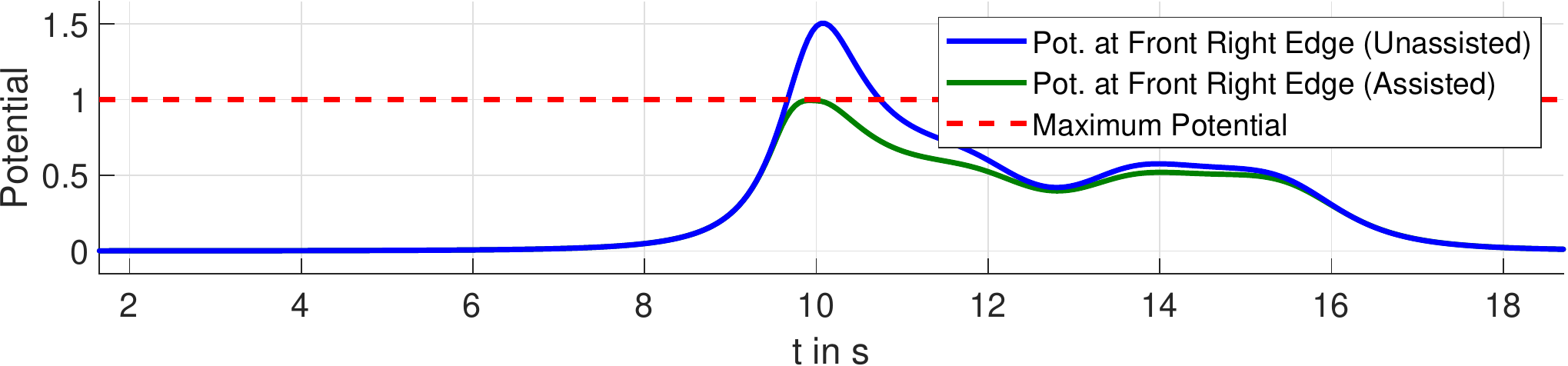}
    \label{fig:scenLCpotf}
}
\centering
\caption{Steering Profiles and Potential Field of the Lane Change Scenario}
\label{fig:scenLCsteerNpotf}
\vspace{-3.5mm}
\end{figure}
\subsection{Lane Change Scenario}
In the second scenario, the teleoperator has to avoid two vehicles that are parked in the lane of the controlled vehicle. As shown in Fig.~\ref{fig:scenLCtraj}, he does so by performing a lane change to the left. The traces of the vehicle front edges are shown for the unassisted and assisted run of the simulation. In this scenario, the operator misjudges the timing of the lane change. Without assistance, this leads to the right front edge entering the bounding ellipse of the first obstacle. In the assisted run, the controller corrects the teleoperator's input such that the vehicle keeps a safe distance. \\
The steering profiles of both runs are shown in Fig.~\ref{fig:scenLCsteer}. The potential field, evaluated at the right front edge of the vehicle, is shown in Fig.~\ref{fig:scenLCpotf}. As compared to the unassisted run, the safety constraint of the maximum potential is not violated with the steering corrections of the controller. In this scenario, the feedback-linearized controller gains of the simulated operator were set to correct both, the heading and the lateral error. Consequently, the vehicle ends up in the center of the left lane in both runs.
\newpage
\section{CONCLUSION AND FUTURE WORK}
\label{CH:conclSec5}
This paper presented a predictive steering control approach for semi-autonomous, teleoperated road vehicles. The primary objective is to track the steering reference generated by a human teleoperator. The controller uses potential fields to rectify steering inputs in cases with the risk of collisions. Furthermore, high-order ellipses are integrated into the design as a means of accurately modeling the buffer zones required for avoiding rectangular obstacles in tight driving scenarios. Validation of the control scheme was performed using a simulated teleoperator in the control loop. However, future validation will make use of inputs from an actual human teleoperator. Further work will focus on extending the approach to overcome limitations related to collisions which are only avoidable through braking. Ultimately, this research aims to develop a fully comprehensive safety concept for teleoperated driving, capable of withstanding a multitude of worst-case circumstances, such as loss of connection. \\
\vspace{3mm}
\section*{APPENDIX}
The controller gains and weights in the presented scenarios were set as follows. \\
\begin{table}[h]
\vspace{-5mm}
\begin{center}
\begin{tabular}{|c||c|c|c|c|c|c|}
\hline
& $w_{\delta_r}$ & $w_P$ & $w_{\dot{\delta}}$ & $\gamma_1$ & $\gamma_2$ & $\gamma_3$ \\ \hline
Parking Lot Scenario & $500$ & $0.15$ & $200$ & $0$ & $0.75$ & $0.25$ \\ \hline
Lane Change Scenario & $500$ & $0.15$ & $200$ & $0.5$ & $1.25$ & $0.25$ \\ \hline
\end{tabular}
\end{center}
\end{table}
\section*{CONTRIBUTIONS AND ACKNOWLEDGMENTS}
Andreas Schimpe was the initiator of the research idea. Frank Diermeyer made essential contributions to the conception of the research project. The research was conducted with basic research funds from the Institute for Automotive Technology. 
\end{document}